\pdfoutput=1

\documentclass[11pt]{article}

\usepackage[final]{acl}

\usepackage{times}
\usepackage{latexsym}

\usepackage[T1]{fontenc}

\usepackage[utf8]{inputenc}

\usepackage{microtype}

\usepackage{inconsolata}

\usepackage{graphicx}
\usepackage{booktabs}
\usepackage{subcaption}
\usepackage{xcolor}   
\usepackage{colortbl} 
\usepackage{makecell}

\usepackage{microtype}
\usepackage{multirow}
\usepackage{caption}
\usepackage{tikz}
\usepackage{amsmath}
\usepackage{cleveref}
\usepackage{enumitem}
\usepackage{courier}

\definecolor{lightergray}{gray}{0.9} 
\definecolor{posgreen}{HTML}{3C763D}
\definecolor{negred} {HTML}{A94442}
\newcommand{\rev}{\textcolor{black}}
\makeatletter

\newcommand*{\@rowstyle}{}

\newcommand*{\rowstyle}[1]{%
  \gdef\@rowstyle{#1}%
  \@rowstyle\ignorespaces%
}

\newcolumntype{=}{%
  >{\gdef\@rowstyle{}}%
}

\newcolumntype{+}{%
  >{\@rowstyle}%
}

\newcommand{\tagC}{\textsuperscript{$\ast$}}
\newcommand{\tagS}{\textsuperscript{\dag}}   
\newcommand{\tagU}{\textsuperscript{\ddag}}      

\makeatother

\title{
    Unsupervised Hallucination Detection by Inspecting Reasoning Processes
}

\author{
    Ponhvoan Srey \quad
    Xiaobao Wu\thanks{Corresponding Authors.} \quad
    Anh Tuan Luu$^{*}$ \\
    Nanyang Technological University \\
    \texttt{\{ponhvoan002, xiaobao002, anhtuan.luu\}@ntu.edu.sg}
}

\begin{document}
\maketitle
\begin{abstract}

    Unsupervised hallucination detection aims to identify hallucinated content generated by large language models (LLMs) without relying on labeled data. 
    While unsupervised methods have gained popularity by eliminating labor-intensive human annotations, they frequently rely on proxy signals unrelated to factual correctness. 
    This misalignment biases detection probes toward superficial or non-truth-related aspects, limiting generalizability across datasets and scenarios. 
    To overcome these limitations, we propose IRIS, an unsupervised hallucination detection framework, leveraging internal representations intrinsic to factual correctness. %
    IRIS prompts the LLM to carefully verify the truthfulness of a given statement, and obtain its contextualized embedding as informative features for training. 
    Meanwhile, the uncertainty of each response is considered a soft pseudolabel for truthfulness. 
    Experimental results demonstrate that IRIS consistently outperforms existing unsupervised methods. 
    Our approach is fully unsupervised, computationally low cost, and works well even with few training data, making it suitable for real-time detection.~\footnote{The code repository is available online at \url{https://github.com/ponhvoan/iris}.}

\end{abstract}

\section{Introduction}

In recent years, Large Language Models (LLMs) such as GPT-4 \citep{achiam2023gpt} have demonstrated remarkable capabilities to generate coherent and relevant responses to user queries. Their success led to the broad adoption of LLMs in a wide variety of tasks, from coding to text summarization \citep{touvron2023llama2, yang2024qwen2, driess2023palm}. However, a critical concern arises because of their tendency to hallucinate, which refers to the phenomenon where LLMs generate texts that appear logical and coherent but contain fictitious information \citep{zhang2023siren}. As LLMs have been observed to confidently generate false information, it is challenging for users to identify hallucinations. This represents a significant drawback to the robustness of LLMs in real-world applications, resulting in growing interest in hallucination detection.

\begin{figure}[!t]
    \centering
    \includegraphics[width=\linewidth]{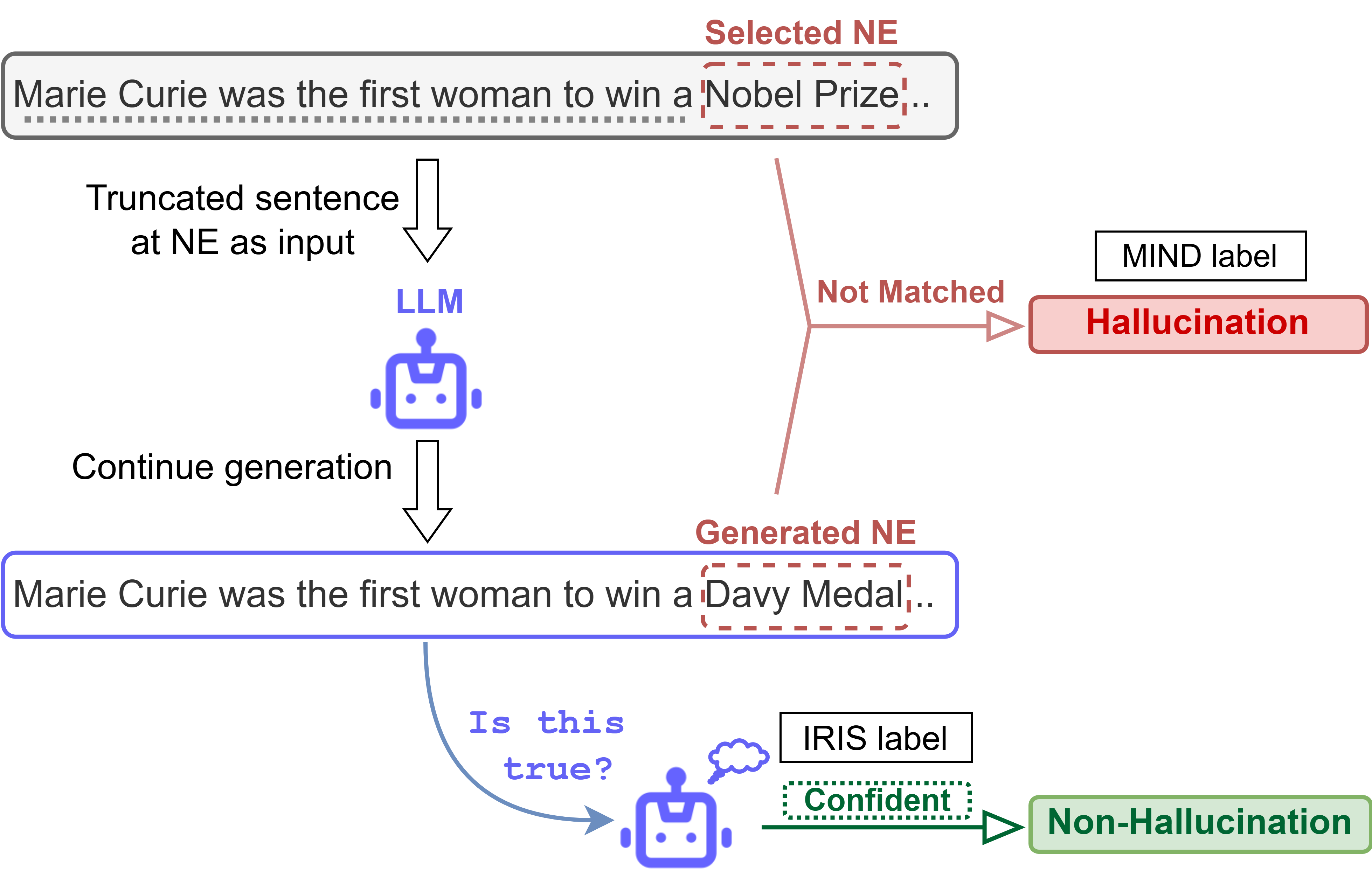}
    \caption{
    Comparison between MIND \citep{su2024unsupervised} and our method.
    MIND incorrectly identifies a statement as hallucination.
    Our method extracts model internal knowledge by asking it to think carefully whether the statement is correct.
    Its confidence is obtained as a soft pseudolabel.}
    \label{fig:label}
\end{figure}

Current hallucination detection strategies are concerned with determining the truth or falsity of a generated statement \citep{su2024unsupervised, manakul2023selfcheckgpt, min2023factscore}. 
A straightforward paradigm is to directly instruct a language model to determine the truth or falsehood of a given text \citep{li-etal-2024-dawn, su2024unsupervised}. Others scrutinize the internal activations for latent signals that may correlate with factual accuracy \citep{su2024unsupervised, azaria2023internal, burns2022discovering}. On the other hand, other methods regard the model uncertainty during text generation as a proxy for hallucination risk.
Generally, state-of-the-arts methods aim to extract the inherent knowledge of LLMs to verify the veracity of statements, whether by directly querying the model, or by inspecting its internal states, or by measuring its uncertainty.

Unfortunately, these methods remain suboptimal for practical deployment. First, they are expensive and incur high computational overhead. Direct querying methods achieve satisfactory performance primarily with high-capacity commercial models, such as GPT-4 \citep{li-etal-2024-dawn, azaria2023internal}, rendering them impractical for open-source alternatives and cost-sensitive applications. Uncertainty-based techniques are hampered by ambiguous threshold determination and often require generating multiple samples per query to reach acceptable performance \citep{kuhn2023semantic, fadeeva2023lmpolygraph, chen2024inside}. Similarly, internal activation approaches depend heavily on extensive manual annotations to train a probe.
Second, methods that bypass the high computational demands risk introducing bias into the detection process. \citet{burns2022discovering} and \citet{su2024unsupervised} consider model internal activations under the unsupervised scenario. Such methods are limited in scope and do not inherently capture the notion of truth, biasing the probe towards non-truth-related signals. 
These limitations highlight critical gaps in current methodologies and motivate the need for more robust, scalable, and efficient frameworks for real-time hallucination detection. A key challenge is to obtain truth labels to train a classifier probe based on model internal states without resorting to human annotations.

In this work, we propose \textbf{Internal Reasoning for Inference of Statement veracity} (\textbf{IRIS}), an unsupervised hallucination detection method that inspects the uncertainty of the reasoning process.
We assert that the model's confidence or uncertainty when verifying the truthfulness of a statement reveals the likelihood of hallucination. This uncertainty is regarded as a soft pseudolabel for statement accuracy. 
To illustrate, let us consider the statement ``Marie Curie was the first woman to win a Davy Medal'', which is a continuation of a truncated Wikipedia sentence (see \Cref{fig:label}). 
As Marie Curie is a well-known public figure, LLMs would retain knowledge that she is also indeed the first woman to win a Davy Medal. When asked to assess the statement, LLMs would exhibit high confidence that it is correct, thus most likely a non-hallucination. IRIS regards this confidence score as a soft pseudolabel for truth.
On the other hand, in MIND \citep{su2024unsupervised}, the label is derived by matching the original and generated NE, incorrectly resulting in ``Hallucination''. MIND biases the labels towards matching named entities, rather than the truth value of the statement.

Furthermore, the model internal states when thinking about the statement represent its latent knowledge, and contain information indicative of factual correctness.
IRIS trains a lightweight probe on the contextualized embeddings of the model verification with the pseudolabels.
Our advantage is that we only need one query to the model for each statement whereas uncertainty-based methods demand multiple samples.

IRIS alleviates the burden of human annotations and promotes real-time hallucination detection. To summarize, the contributions of this paper are as follows:
\begin{itemize}[leftmargin=*]
    \item We propose an unsupervised hallucination detection framework where the pseudolabels are related to the truthfulness of the statements and the features are model internal states.
    \item We demonstrate that the contextualized embeddings of the model's verification is more informative of the hallucination risk, compared to those of the statements themselves.
    \item We conduct extensive experiments and show that our method achieves an improvement of 3.2\%, 7.0\%, and 10.2\% on unsupervised hallucination detection with the True-False, HaluEval2, and HELM datasets, respectively.
\end{itemize}

\begin{figure*}[ht]
    \includegraphics[width=\textwidth]{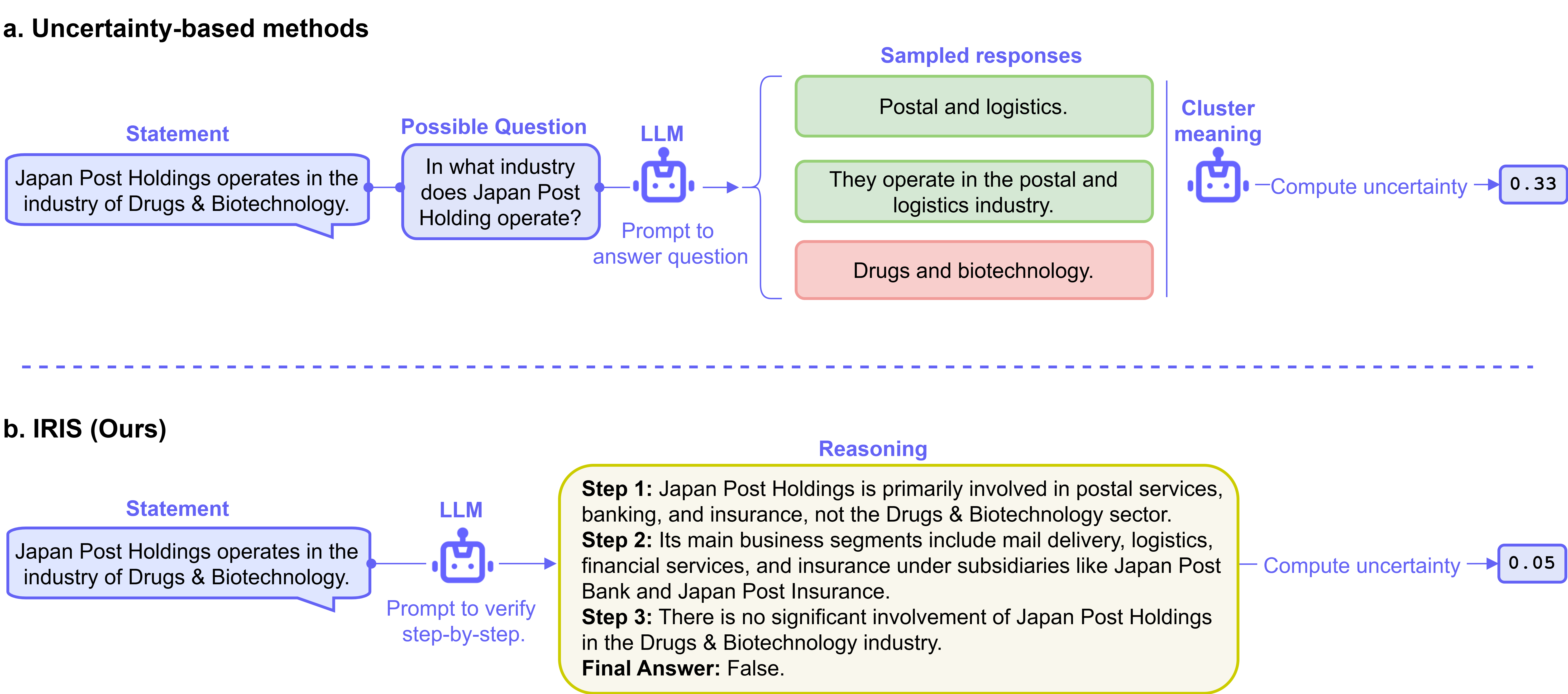}
    \caption{
    Illustrations of eliciting uncertainty.
    In uncertainty-based methods, given a statement, a question to which the statement could have been the answer is generated. Multiple responses from the LLM are then sampled, and the responses are aggregated based on meaning and the uncertainty is computed. Meanwhile, our approach only calls the LLM \emph{once} using CoT prompting to facilitate a more thoughtful and informative response. The uncertainty is normalized so that 0 indicates hallucination.}
    \label{fig:diag}
\end{figure*}

\section{Methodology}

In this section, we present the details of our proposed unsupervised hallucination method based on internal activations. We note that our method is designed to detect hallucinations in LLMs as the primary goal, but our proposal is general and is thus applicable to identifying falsehoods by leveraging the internal knowledge of LLMs.

\subsection{Eliciting Internal Knowledge}

Given a set $(s_1, \ldots, s_n)$ of statements, our goal is to determine whether each $s_i$ is true. Here, we allow for the most general case where each statement could be either human-written or generated by an external language model that we no longer have access to. Previous research \cite{azaria2023internal, ji2024llm} show that the LLMs contain internal knowledge, accessed via their internal states, on the truthfulness of statements. 
Additionally, we believe that the contextualized embeddings of the response verifying the statements is a more reliable predictor of hallucination than those of the statements themselves. 
As such, we first prompt the model to carefully verify whether a statement is correct. 
Section~\ref{sec:expt} provides the results of using the embeddings of the original embeddings (SAPLMA) and of the verification response.

\subsection{Uncertainty as Pseudolabels}

Importantly, we require that our method does not rely on external supervision. To this end, we derive a pseudolabel for each statement. Ultimately, we wish to train a classifier probe on the internal states and these pseudolabels. To ensure that the classifier is endowed with the capacity to tell truth from falsehood, these pseudolabels should encompass some notion of truth. 

In uncertainty-based hallucination detection, \rev{e.g. as formulated by \citet{kuhn2023semantic} and \citet{duan-etal-2024-shifting}}, the underpinning assumption is that when an LLM is responding to a query, its level of uncertainty indicates its lack of knowledge and the likelihood of hallucinating. 
As the uncertainty in the response directly links to hallucination risk and captures some aspect of factuality, we want to leverage the same insight and make use of uncertainty as pseudolabels.
However, these methods involve passing the query to the model and evaluating its output. This imposes three key demands. First, to measure an LLM's uncertainty regarding a statement, a question to which the statement might be the answer is generated either using the same LLM or a different one. However, this may propagate errors as the question may be vague and lead to answers inconsistent with the provided statement. Second, multiple calls to the LLM is required to sample responses. Lastly, an additional model is used to aggregate the responses according to semantic meanings to estimate uncertainty. Overall, this approach is compute heavy and less practical for real-time detection. In our case, we seek to bypass these requirements. 

With this goal in mind, given a statement $s_i$, we prompt an LLM to evaluate its correctness. We argue that the uncertainty in its appraisal equates to the model's knowledge regarding the statement and indicates the truthfulness of the sentence. Further, we prompt the LLM to reason step-by-step, eliciting a more thoughtful and informative response as the model assesses the correctness of the statements. As illustrated in \Cref{fig:diag}, our approach involves calling the LLM only once, and the reasoning chain is examined to estimate the LLM's uncertainty regarding the provided statement.

\begin{figure}[!t]
    \centering
    \includegraphics[width=\linewidth]{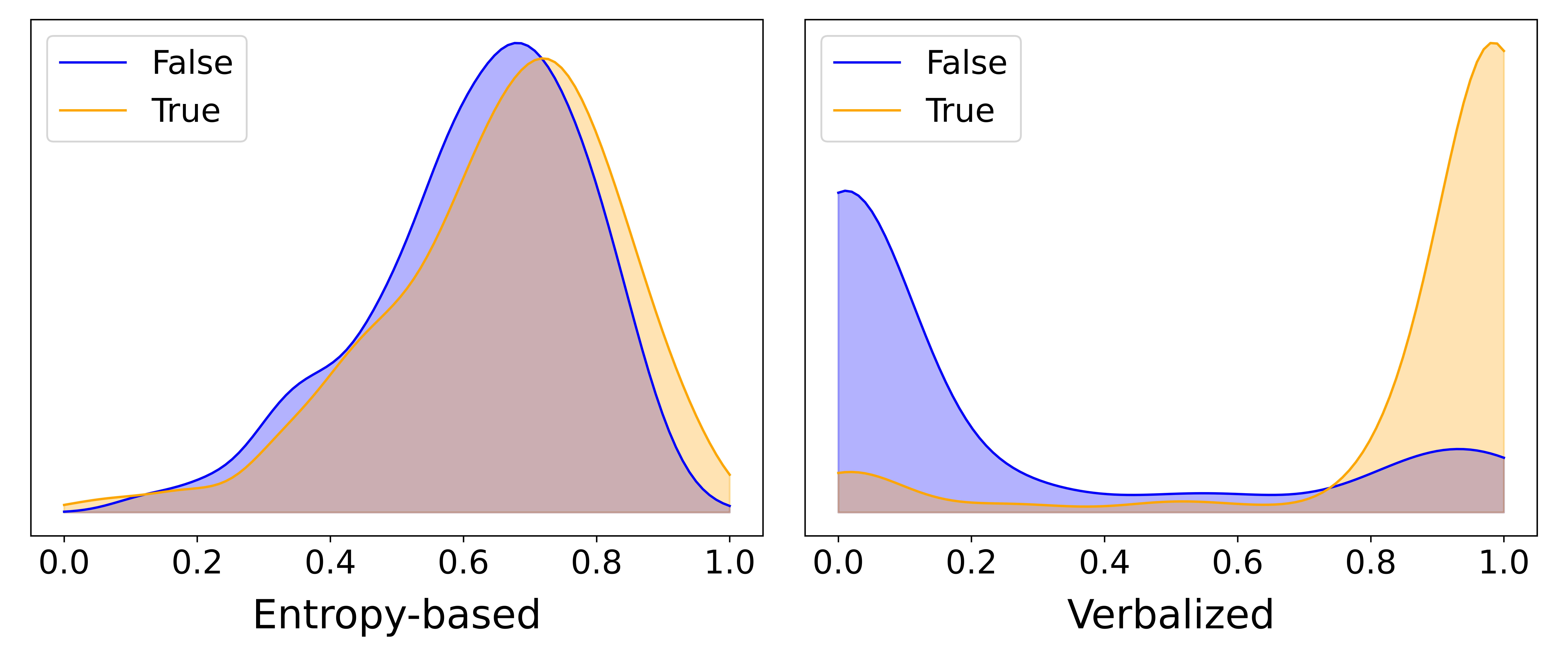}
    \caption{Distribution of confidence in determining the correctness of statements from the True-False dataset.}
    \label{fig:ent_v_verb}
\end{figure}

To estimate the uncertainty, we explore two options: (a). entropy-based: calculate the entropy of the reasoning chain using token probabilities, and normalize the entropy such that a value of 1 indicates correctness; and (b). verbalized: the LLM expresses the confidence that the statement is correct in numerical probabilities in the context of the preceding reasoning steps. The prompt templates are provided in Appendix~\ref{app:prompt}. Similar to \rev{the findings by \citet{tian2023just}}, we find that the verbalized confidence is better calibrated compared to token probabilities as shown in \Cref{fig:ent_v_verb}.

\subsection{Classifier Training}

Suppose the evaluation of the statement $s_i$ yields a response $x_i$, we obtain the contextualized embeddings $\phi(x_i)$ from the LLM.
In line with SAPLAMA \citep{azaria2023internal} and MIND \citep{su2024unsupervised}, we utilize the last token's embeddings at the last layer. In \Cref{subsec:analysis}, we study the efficacy of using embeddings at other layer depths.

For each $\phi(x_i)$, we generate a soft pseduolabel $\tilde{y}_i \in [0,1]$, which is the uncertainty of the LLM's assessment of whether the statement $s_i$ is true. We normalize $\tilde{y_i}$ so that $\tilde{y_i}=1$ means $s_i$ is correct, and vice versa. 
To address the issue of noisy labels, we apply soft bootstrapping \citep{reed2014training}, where new regression targets are generated for each mini-batch based on the classifier's current predictions. The updated target for the $i$-th statement is $t_i = \beta \tilde{y_i} + (1-\beta) \hat{y}_i$, where $q_i$ is the classifier's prediction, and $\beta \in (0,1)$. Further, we employ the symmetric cross entropy loss \citep{wang2019symmetric} as the overall objective
\begin{equation}
    l_i = l_{ce} + l_{rce} = H(\hat{y}_i, t_i) + \phi H(t_i, \hat{y}_i),
\end{equation}

\noindent where $H(\hat{y}_i, t_i) = t_i \log \hat{y}_i + (1 - t_i) \log (1 - \hat{y}_i)$ is the standard cross entropy loss, $H(t_i, \hat{y}_i)$ is the reverse cross entropy loss, and $\phi$ is a hyperparameter balancing the losses. The cross entropy term aligns the predictions with the pseudolabels, but this causes the classifier to be sensitive to noisy labels. On the other hand, the reverse cross entropy term penalizes the classifier for being too confident on potentially incorrect targets. Their combination prevents overfitting to noisy labels, and allows for better generalization. 

To reduce computational demands, the classifier is implemented as a small feedforward MLP with three fully-connected hidden layers of units $(256, 128, 64)$, each followed by ReLU activation. In the final layer, a sigmoid activation is applied. The Adam optimizer is used for training, with a learning rate of $10^{-2}$, and weight decay of $10^{-5}$. 
We present hyperparameter finetuning results in Appendix~\ref{sec:ablation}. 
For all settings, the classifier is trained for 10 epochs with a patience of 5 epochs. 
In Appendix~\ref{app:sensitivity}, results on prompt sensitivity are reported.

\section{Experiments}
\label{sec:expt}

\begin{table*}[hbt!]
    \centering
    \setlength{\tabcolsep}{5.5mm}
    \renewcommand{\arraystretch}{1.2}
    \begin{subtable}{\textwidth}
        \centering
        \resizebox{\linewidth}{!}{
        \begin{tabular}{=l|+c+c+c+c+c+c+c|+c} 
            \toprule
            \textbf{Method} & \textbf{Animals} & \textbf{Cities} & \textbf{Companies} & \textbf{Elements} & \textbf{Facts} & \textbf{Generation} & \textbf{Invention} & \textbf{Average}  \\
            \midrule
            \rowstyle{\color{black}}
            Zero-shot\tagU  & 80.06 & 91.34 & 90.33 & 81.83 & 92.17 & 79.18 & 75.46 & 85.37 \\
            Few-shot\tagU   & 78.08 & 89.23 & 89.75 & 82.90 & 92.00 & 75.51 & 89.38 & 86.38  \\
            CoT (zero-shot)\tagU & \textbf{80.65} & 92.04 & 91.67 & 86.88 & 92.82 & 83.27 & 80.48 & 87.54 \\            
            CoT (few-shot)\tagU & 75.69 & 92.39 & 91.00 & 86.34 & 91.19 & 78.78 & 83.11 & 86.65 \\
            \rowstyle{\color{black}} 
            EigenScore\tagU & 57.45 & 63.04 & 53.57 & 51.15 & 51.28 & 60.23 & 53.67 & 56.06 \\
            \rowstyle{\color{black}} 
            SAR\tagU & 64.68 & 58.04 & 62.74 & 53.15 & 58.28 & 70.76 & 58.56 & 59.86 \\
            CCS\tagU        & 67.33 & 51.71 & 57.08 & 70.97 & 79.67 & 57.14 & 67.61 & 63.16 \\
            MIND\tagU       & 51.49 & 53.08 & 53.75 & 44.62 & 54.47 & 36.73 & 61.36 & 52.36 \\
            \textbf{IRIS\tagU (ours)}       & 80.20 & \textbf{93.84} & \textbf{94.17} & \textbf{89.25} &  \textbf{93.50} & \textbf{87.76} & \textbf{90.91} & \textbf{90.38} \\
            \midrule
            GPT-4o\tagC & 82.74 & 94.10 & 91.42 & 96.02 & 98.53 & 82.45 & 86.30 & 90.96 \\
            SAPLMA\tagS    & 84.16 & 95.21 & 87.92 & 84.41 & 93.50 & 79.59 & 94.89 & 89.67 \\
            Ceiling\tagS & 81.19 & 95.21 & 94.17 & 91.40 & 95.21 & 87.76 & 90.34 & 91.25 \\
            \bottomrule
        \end{tabular}
        }
        \caption{True-False}
        \label{tab:tf_acc}
    \end{subtable}
    
    \begin{subtable}{\textwidth}
        \centering
        \setlength{\tabcolsep}{7mm}
        \renewcommand{\arraystretch}{1.2}
        \resizebox{\linewidth}{!}{
        \begin{tabular}{=l|+c+c+c+c+c|+c} 
            \toprule
            \textbf{Method} & \textbf{Bio-Medical} & \textbf{Education} & \textbf{Finance} & \textbf{Open-Domain} & \textbf{Science} & \textbf{Average} \\
            \midrule
            \rowstyle{\color{black}} 
            Zero-shot\tagU  & 59.90 & 63.48 & 66.53 & 73.81 & 61.25 & 63.67 \\
            Few-shot\tagU  & 58.33 & 59.23 & 71.93 & 55.78 & 59.04 & 61.88 \\
            CoT (zero-shot)\tagU& 56.28 & 61.29 & 67.58 & 63.61 & 56.83 & 60.77 \\
            CoT (few-shot)\tagU & 59.18 & 62.11 & 58.05 & 63.95 & 55.52 & 58.87 \\
            \rowstyle{\color{black}} 
            EigenScore\tagU & 41.80 & 46.58 & 45.15 & 58.05 & 35.01 & 43.03 \\
            \rowstyle{\color{black}} 
            SAR\tagU & \textbf{62.35} & 63.01 & 63.94 & 57.56 & \textbf{75.61} & 65.98 \\
            CCS\tagU        & 53.61 & 60.54 & 61.94 & 54.24 & 53.00 & 56.91 \\
            MIND\tagU       & 40.36 & 45.58 & 75.66 & 30.51 & 45.50 & 50.74 \\
            \textbf{IRIS\tagU (ours)}       & 59.64 & \textbf{63.95} & \textbf{78.84} & \textbf{93.22} & 70.00 & \textbf{70.57} \\
            \midrule
            GPT-4o\tagC & 71.38 & 83.47 & 71.18 & 73.87 & 81.29 & 75.59 \\
            SAPLMA\tagS    & 59.64 & 64.63 & 77.78 & 93.22 & 81.00 & 73.33 \\
            Ceiling\tagS & 72.89 & 67.35 & 77.25 & 93.22 & 81.50 & 76.74 \\
            \bottomrule
        \end{tabular}
        }
        \caption{HaluEval2}
        \label{tab:halueval_acc}
    \end{subtable}

    \begin{subtable}{\textwidth}
        \centering
        \setlength{\tabcolsep}{6.5mm}
        \renewcommand{\arraystretch}{1.2}
        \resizebox{\linewidth}{!}{
        \begin{tabular}{=l|+c+c+c+c+c+c|+c} 
            \toprule
            \textbf{Method} & \textbf{Falcon} & \textbf{GPT-J} & \textbf{LLB-7B} & \textbf{LLC-7B} & \textbf{LLC-13B} & \textbf{OPT-7B} & \textbf{Average} \\
            \midrule
            \rowstyle{\color{black}} 
            Zero-shot\tagU  & 56.81 & 64.51 & 57.17 & 56.34 & 61.34 & 64.49 & 60.21 \\
            Few-shot\tagU   & 58.73 & 66.43 & 60.88 & 48.09 & 44.12 & 64.66 & 57.71 \\
            CoT (zero-shot)\tagU& 62.57 & 68.71 & 58.94 & 58.35 & 57.35 & 65.55 & 62.12 \\
            CoT (few-shot)\tagU& 56.62 & 64.86 & 60.53 & 54.53 & 56.09 & 66.43 & 60.12 \\
            \rowstyle{\color{black}} 
            EigenScore\tagU & 38.46 & 42.25 & 44.30 & 37.18 & 42.34 & 48.99 & 42.41 \\
            \rowstyle{\color{black}} 
            SAR\tagU & 61.26 & 57.50 & 64.30 & 59.94 & 57.66 & 51.52 & 58.66 \\
            CCS\tagU & 55.24 & 50.43 & 52.21 & 57.00 & 61.46 & 64.04 & 56.79 \\
            MIND\tagU & 54.29 & 52.17 & 59.29 & 44.00 & 59.38 & 56.14 & 54.10 \\
            \textbf{IRIS\tagU (ours)}       & \textbf{67.62} & \textbf{69.57} & \textbf{69.03} & \textbf{66.00} & \textbf{69.79} & \textbf{68.42} & \textbf{68.43} \\
            \midrule
            GPT-4o\tagC & 70.25 & 76.57 & 70.97 & 57.34 & 62.39 & 77.56 & 69.63 \\
            SAPLMA\tagS    & 73.33 & 85.22 & 56.64 & 76.00 & 68.75 & 81.58 & 73.61 \\
            Ceiling\tagS & 77.14 & 83.48 & 66.37 & 85.00 & 71.88 & 82.46 & 77.80 \\ 
            \bottomrule
        \end{tabular}
        }
        \caption{HELM}
        \label{tab:helm_acc}
    \end{subtable}
    \caption{
    Accuracy results on the True-False, HaluEval2, and HELM datasets with three kinds of methods: Unsupervised (\tagU), Supervised (\tagS), and Commercial (\tagC).
    The best results of unsupervised methods are in \textbf{bold}.
    }
    \label{tab:main_results}
\end{table*}

\subsection{Datasets} 

We evaluate our method on three recent hallucination datasets: (i) \textbf{True-False} dataset \citep{azaria2023internal}: a compilation of $\sim$6300 factual statements across 6 topics (Animals, Cities, Companies, Elements, Facts, and Inventions) and a set of LLM generated statements (Generated);
(ii) a subset of \textbf{HaluEval2} with human annotations \citep{li-etal-2024-dawn}, which consists of a total of $\sim$3800 ChatGPT responses to challenging queries related to Bio-Medical, Education, Finance, Open-Domain, and Science;
(iii) \textbf{HELM} \citep{su2024unsupervised}: a collection of  $\sim$3600 LLM-generated text continuation based on Wikipedia articles across six LLMs of varying sizes.
Together, these datasets represent a balanced mix of LLM-generated and non-LLM statements, covering a wide range of topics. The topic or sub-dataset in each dataset is split 80-20 for training and testing.

\subsection{Baselines} 

In the main experiments, we use \textbf{Llama-3.1-8B-Instruct} \citep{dubey2024llama} as the proxy model to extract knowledge from and to assess the correctness of statements. 
We evaluate our proposal against three different types of baselines. 

The first is direct prompting, where the LLM is asked to determine whether a statement is factual under the zero-shot and few-shot (three demonstrations) settings. We utilize two prompting techniques: asking the model to directly determine correctness, and chain-of-thought (CoT) prompting. This results in \textbf{Zero-shot}, \textbf{Few-shot}, \textbf{CoT (zero-shot)}, and \textbf{CoT (few-shot)}. To compare to commercial LLMs, following the approach in HaluEval \citep{li-etal-2023-halueval}, results from querying \textbf{GPT-4o} are provided.

\rev{Secondly, we evaluate the datasets with two recent uncertainty-based methods, \textbf{EigenScore} \cite{chen2024inside} and \textbf{SAR} \citep{duan-etal-2024-shifting}. For each statement, the LLM is queried 10 times to determine its confidence of the statement correctness, and an uncertainty score is computed from these samples. To follow the unsupervised setting, sentences with uncertainty scores greater than the median for each dataset are considered incorrect.}

Next, we compare with methods most closely related to our work --- unsupervised hallucination detection based on LLM internal activations. In particular, we compare with Contrast-Consistent Search (\textbf{CCS}) \citep{burns2022discovering} and \textbf{MIND} \citep{su2024unsupervised}. CCS converts every sentence into a pair of contrasting statements, and trains a probe such that the probabilities of each pair being true (the statement itself and its negation) adds up to one. MIND truncates Wikipedia articles and prompts an LLM to continue the next sentence. The truth labels are automatically generated by matching the named entities of the response and of the succeeding sentence in the original article. MIND assumes access to the LLMs that were used to generate the statements and use their contextualized embeddings as the features for training the probe. To conform to the same setting as our method, we do not have such access, and instead make use of the proxy model. Note that MIND's probe is trained on its own automatically labeled dataset before testing on the validation sets.

Finally, we provide results of supervised baselines for comparison. \textbf{SAPLMA} \citep{azaria2023internal} utilizes the contextualized embeddings of the statements, and trains a probe using these and ground-truth labels. 
Lastly, the supervised \textbf{Ceiling} trains the probe on the embeddings of the model's statement verification and the ground-truth labels. All classifier-based methods employ the same MLP architecture.

Since the main task is binary classification, detection accuracy is used as the evaluation metric. For probe-based methods, the validation accuracy on a held-out set is computed.

\section{Experimental Results and Analysis}
\subsection{Main Results}
\Cref{tab:main_results} presents the main results across all sub-topics of the True-False, HaluEval2, and HELM datasets. 
Among all unsupervised methods using Llama-3.1-8B-Instruct, our method attains the highest average accuracy for all datasets, outperforming the best baseline by 3.2\%, 7.0\%, and 10.2\% on the True-False, HaluEval2, and HELM datasets, respectively. Overall, our method is the most consistent, obtaining the best accuracy for 15 out of 18 sub-datasets. 
Although the commercial GPT-4o performs better, IRIS uses a much smaller 8B model, and is able to achieve comparable accuracy on the True-False and HELM datasets.

While Chain-of-Thought may help improve accuracy, it is not consistent. This shows that direct query as a method to detect hallucination is sensitive to the input prompt. Meanwhile, the unsupervised detection methods, CCS and MIND, mostly perform worse than direct query. In particular, MIND reached a training and validation accuracy of approximately 76\% and 70\% on their automatically annotated data. However, its accuracy drops significantly when tested on other datasets, indicating its inability to generalize well. 
Notably, the supervised ceiling exceeds the performance of SAPLMA. Recall that SAPLMA utilizes the embeddings of the statements themselves as the training features whereas the supervised ceiling uses those of the model verification process of the statements. These results reinforce our hypothesis that the latter is more informative than the former.

\subsection{Analysis}
\label{subsec:analysis}

In this section, we analyze IRIS to evaluate its performance under various scenarios.

\begin{table}[!t]
  \centering
  \setlength{\tabcolsep}{1mm}
  \renewcommand{\arraystretch}{1}
  \resizebox{\linewidth}{!}{
  \begin{tabular}{@{} l| *{6}{c} @{}}
    \toprule
    \multirow{2}{*}{\textbf{Train on}} & \multicolumn{6}{c}{\textbf{Test on}} \\
    \cmidrule(l){2-7}
    & Animals & Cities & Bio-Med & Education & Falcon & GPT-J \\
    \midrule
    Animals & 80.20 & \makecell[c]{91.78\\\small\textcolor{negred}{(-2.06)}} & \makecell[c]{60.84\\\small\textcolor{posgreen}{(+1.20)}} & \makecell[c]{67.35\\\small\textcolor{posgreen}{(+3.40)}} & \makecell[c]{63.81\\\small\textcolor{negred}{(-3.81)}} & \makecell[c]{68.70\\\small\textcolor{negred}{(-0.87)}} \\
    \midrule 
    Cities 
      & \makecell[c]{73.76\\\small\textcolor{negred}{(-6.44)}} & 93.84 & \makecell[c]{57.83\\\small\textcolor{negred}{(-1.81)}} & \makecell[c]{68.03\\\small\textcolor{posgreen}{(+4.08)}} & \makecell[c]{57.14\\\small\textcolor{negred}{(-10.48)}} & \makecell[c]{65.22\\\small\textcolor{negred}{(-4.35)}} \\
    \midrule
    Bio-Med  
      & \makecell[c]{74.26\\\small\textcolor{negred}{(-5.94)}} & \makecell[c]{86.99\\\small\textcolor{negred}{(-6.85)}} & 59.64 & \makecell[c]{63.27\\\small\textcolor{negred}{(-0.68)}} & \makecell[c]{62.86\\\small\textcolor{negred}{(-4.76)}} & \makecell[c]{63.48\\\small\textcolor{negred}{(-6.09)}} \\
    \midrule 
    Education
      & \makecell[c]{73.27\\\small\textcolor{negred}{(-6.93)}} & \makecell[c]{89.04\\\small\textcolor{negred}{(-4.80)}} & \makecell[c]{65.66\\\small\textcolor{posgreen}{(+6.02)}} & 63.95 & \makecell[c]{62.86\\\small\textcolor{negred}{(-4.76)}} & \makecell[c]{67.83\\\small\textcolor{negred}{(-1.74)}} \\
    \midrule
    Falcon
      & \makecell[c]{79.70\\\small\textcolor{negred}{(-0.50)}} & \makecell[c]{90.75\\\small\textcolor{negred}{(-3.09)}} & \makecell[c]{59.04\\\small\textcolor{negred}{(-0.60)}} & \makecell[c]{59.18\\\small\textcolor{negred}{(-4.77)}} & 67.62 & \makecell[c]{66.96\\\small\textcolor{negred}{(-2.61)}} \\
    \midrule 
    GPT-J
      & \makecell[c]{74.75\\\small\textcolor{negred}{(-5.45)}} & \makecell[c]{83.90\\\small\textcolor{negred}{(-9.94)}} & \makecell[c]{60.84\\\small\textcolor{posgreen}{(1.20)}} & \makecell[c]{59.86\\\small\textcolor{negred}{(-4.09)}} & \makecell[c]{61.90\\\small\textcolor{negred}{(-5.72)}} & 69.57 \\
    \bottomrule
  \end{tabular}
  }
  \caption{Accuracy of trained probes on out-of-distribution test set. 
  The performance decrease and increase are highlighted in \textcolor{negred}{red} and \textcolor{posgreen}{green}, respectively.}
  \label{fig:ood}
\end{table}

\paragraph{Out-of-Distribution (OOD) Setting.} To test the performance of IRIS under the OOD setting, we consider two topic from each of the True-False, HaluEval2, and HELM datasets. \Cref{fig:ood} reports the OOD performance. On average, the accuracy decreases by 3.1\%, but remains generally robust. With the True-False datasets used for training, the OOD test accuracy decreases slightly by 2.1\%, compared to a drop of 3.7\% and 3.6\% when HaluEval2 and HELM are used. We can observe that when trained on easier datasets where the model has more knowledge, the test accuracy decreases less drastically or even improves slightly. For example, for ``Animals'', the average OOD accuracy only decreases marginally by 0.4\%, and its test accuracy on ``Bio-Medical'' and ``Education'' improves by 1.2\% and 3.4\%.

\begin{figure}[!t]
    \centering
    \includegraphics[width=\linewidth]{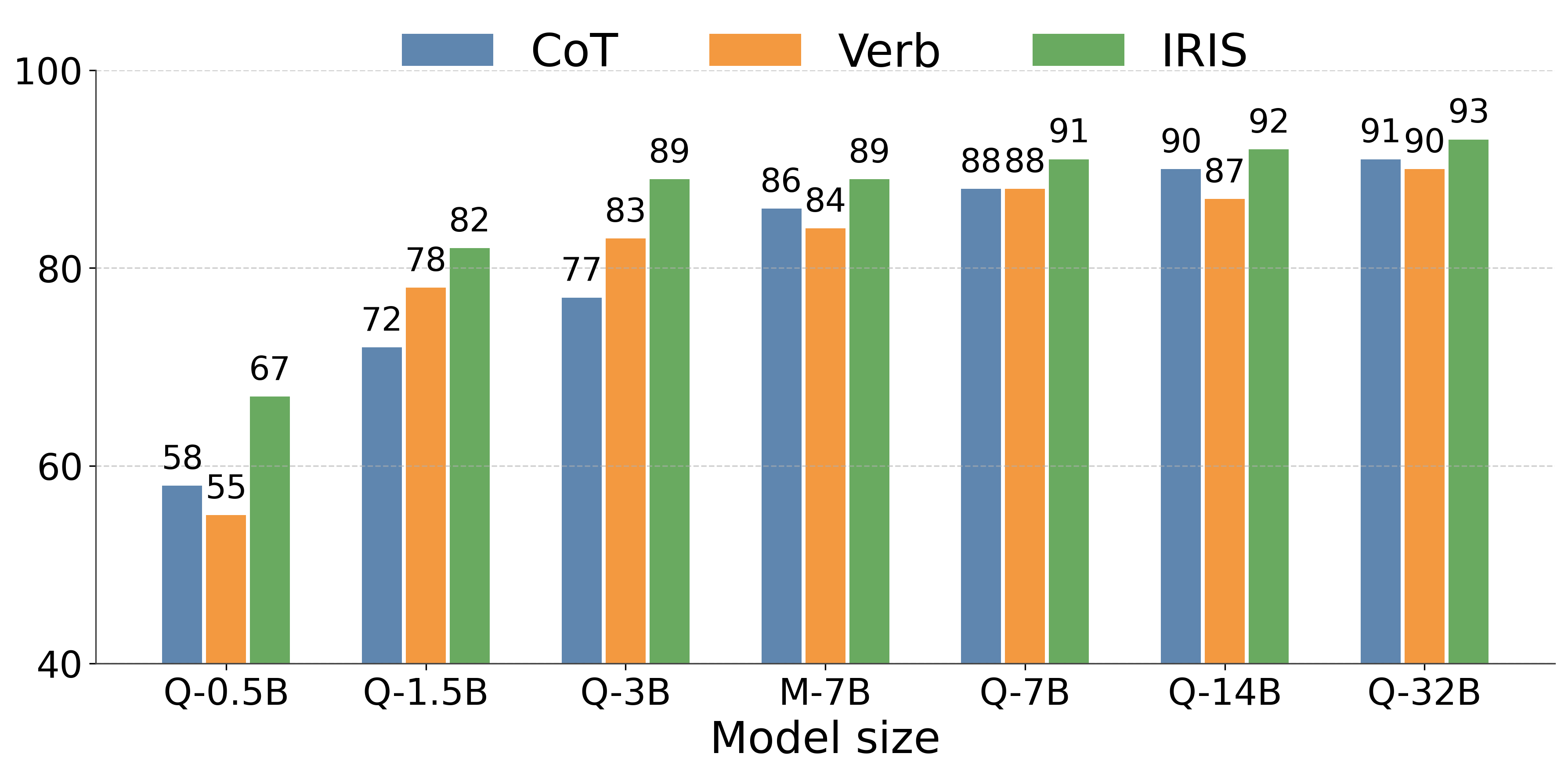}
    \caption{Average accuracy (\%) on True-False dataset with models of different sizes.
    ``Q'' and ``M'' represent instruction-tuned Qwen-2.5 \citep{yang2024qwen2} and Mistral-v0.3 \citep{mistralv0.3} models, respectively.}
    \label{fig:size}
\end{figure}

\paragraph{Model Size.} We test with different model sizes to understand the influence of model capacity and internal knowledge on the performance of IRIS.  \Cref{fig:size} reports the accuracy of IRIS, along with direct prompting using CoT and verbalized confidence (Verb). For Verb, statements with a confidence score above 0.5 are considered factual. 
IRIS consistently performs the best, and the improvement is more significant in smaller models.
With smaller models, the model knowledge regarding factual correctness decreases. The verbalized confidence is not as well-calibrated, and thus, using it to directly identify correct statements yields a significantly worse accuracy. Nonetheless, their hidden states contain richer contextual information compared to the response. Training a probe on the hidden states via the IRIS pipeline leads to greater gains, especially on the smallest 0.5B model where IRIS outperforms Verb by 12\%. 

\begin{figure}
    \centering
    \includegraphics[width=\linewidth]{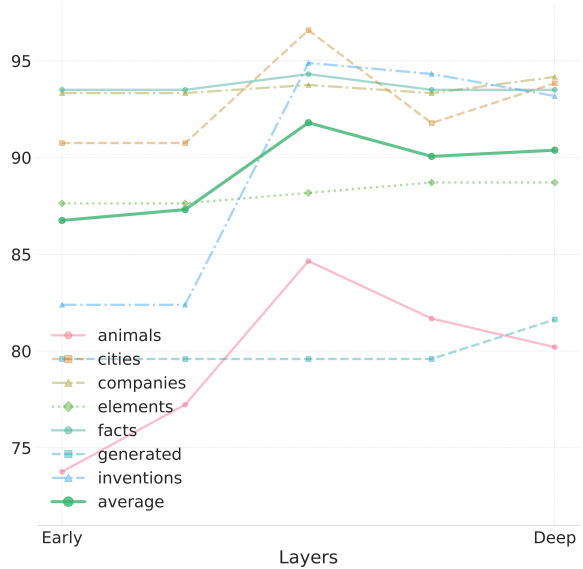}
    \caption{Accuracy (\%) using embeddings at different layers of Llama-3.1-8B-Instruct.}
    \label{fig:layer}
\end{figure}

\paragraph{Layer Depth.} We investigate the contribution of each layer on the overall detection performance. As shown in \Cref{fig:layer}, the embeddings at different layers yield a range of accuracy. The average accuracy peaks with the middle layer, but for some dataset, such as ``Companies'', ``Elements'', and ``Generated'', the last layer gives better results. Our study does not provide conclusive evidence regarding which layer is ideal for hallucination discrimination. In previous works, \citet{ji2024llm} found that the final layer is the best in recognizing falsehoods, whereas \citet{azaria2023internal} advocated for the intermediate layers. We believe that it ultimately depends on the dataset and LLM used. Therefore, we can employ an architecture to integrate the embeddings at all depths to learn better. However, our preliminary consideration of adding a small learnable module to fuse the embeddings did not improve performance. Further examination with more sophisticated architecture is required.

\begin{table}[!t]
    \centering
    \renewcommand{\arraystretch}{1.25}
    \resizebox{0.9\linewidth}{!}{
    \begin{tabular}{c|c c c c c}
         \toprule
         \textbf{\# Data} & 32 & 64 & 128 & 256 & \textbf{All} \\
         \midrule
         \textbf{Acc} & 84.63 & 87.86 & 88.49 & 89.83 & 90.38 \\
         \bottomrule
    \end{tabular}
    }
    \caption{Average True-False accuracy (\%) using different sizes of training data.}
    \label{tab:trsize}
\end{table}

\paragraph{Training Data Size.} IRIS relies on access to an unlabeled collection of statements as training data. In this subsection, we evaluate its performance with training data of different sizes. \Cref{tab:trsize} reports the average True-False accuracy on the validation sets. The experiments indicate that our method improves with more data, but the gains plateau with more than 128 data points. Most importantly, our proposal works reasonably well even with as few as 32 statements. With this observation, given some demonstration statements from HELM, we prompt Llama-3.1-8B-Instruct to generate 32 statements of similar nature and topics, half of which it knows to be true, and the other half false. 
An example generated true statement is ``The largest mammal by mass is the blue whale'', and false statement is ``Google's motto is `Don't be evil'.'' Using the generated statements as training data, the average accuracy on the validation sets is 86.28\%, comparable to when the actual True-False data is used. This implication reduces our reliance on having a large carefully crafted training dataset.

\section{Related Work}

In this section, we provide an overview of existing research on LLM hallucination detection. 
These methods can be broadly categorized into three groups: detection by direct query, by estimating consistency or uncertainty in the response, and by extracting the model's internal knowledge.

\paragraph{Direct Prompting.} \rev{LLMs have been increasingly utilized to evaluate the quality of generated natural language \citep{fu2023gptscore, wang-etal-2023-chatgpt, liu-etal-2023-g}.}
Coupled with the success of few-shot learning \citep{brown2020language}, prompt engineering, a simple strategy to elicit what LLMs know via better prompts, has garnered great attention \citep{sahoo2024systematic}. 
In hallucination detection, improved prompting is often incorporated to more accurately elicit LLMs world knowledge. \citet{li-etal-2024-dawn} instruct GPT-4 \citep{achiam2023gpt} directly to assess factual accuracy given a few demonstrations. \citet{mundler2023self} generate two responses based on the same context and prompt an LLM to evaluate whether there is a contradiction, which implies inaccuracy in one of the responses. Chain-of-Verification (CoVe) \citep{dhuliawala-etal-2024-chain} decomposes a response into individual claims and constructs verification questions for each. For more complex tasks, Chain-of-Thought (CoT) prompting, which requests LLMs to respond to queries by reasoning step-by-step, results in more thoughtful responses \citep{kojima2022large, wei2022chain}. Self-consistent CoT \citep{wang2022self} samples multiple reasoning paths and chooses the most consistent response. Tree of Thoughts (ToT) \citep{yao2023tree} branches out the reasoning chain into intermediate thoughts and allows backtracking and lookahead. An obvious extension is to apply CoT prompting to LLMs to detect cases of hallucination.

\paragraph{Uncertainty Estimation.} Another approach in hallucination detection is to determine the uncertainty of the generated sequence. The fundamental premise is that the higher the uncertainty in the response, the likelier the model has hallucinated \citep{fadeeva2023lmpolygraph}. 
\citet{kuhn2023semantic} propose semantic uncertainty to account for how syntactically different responses that share the same meaning should not increase the uncertainty. Claim-conditioned uncertainty further removes the impact of the uncertainty due to synonyms, different claim types, or order in the answer \citep{fadeeva-etal-2024-fact}. On top of sentence level uncertainty, \citet{duan-etal-2024-shifting} examine uncertainty at the token level, and remove the influence of semantically insignificant tokens.
\citet{zhang2023enhancing} weigh their uncertainty measure with attention values, and address over- and under-confidence issues by correcting the token probability with the inverse document frequency (IDF). 
\citet{chen2024inside} devise EigenScore, an uncertainty metric that measures the spread of the response in the embedding space.
Under the black-box setting, multiple responses are sampled and their consistency is estimated \citep{kuhn2023semantic, manakul2023selfcheckgpt, lin2024generatingconfidence}. 
\rev{However, to estimate an uncertainty score, these approaches generally require multiple calls to the LLM, rendering them computationally heavy.}

\paragraph{Internal Knowledge.} Findings from recent research provide strong evidence that LLMs contain more knowledge about their own response
\cite{pan2024fallacy,wu2023effective,wu2024fastopic,wu2024survey,wu2024akew,wu2024antileak,wu2025sailing}.
\citet{saunders2022self} analyze the generation-discrimination gap, which is the discrepancy between the model's ability to produce high-quality outputs and its capacity to accurately evaluate those outputs. \citet{li2023inference} highlight that attention heads contain crucial information related to factuality, and modify attention activations to enforce more truthful generation. SAPLMA \citep{azaria2023internal} argues that the internal states are discriminative features to uncover statement truthfulness, and train a probe on the internal states with human annotations to label hallucinations. Likewise, \citet{ji2024llm} describe similar findings.
As human annotation is labor-intensive, some studies focus on the unsupervised setting.
\citet{burns2022discovering} train a linear probe through unsupervised clustering by maximizing the distance between the embeddings of a statement and its negation. Meanwhile, MIND \citep{su2024unsupervised} designs a training dataset that is automatically annotated by matching named entities in Wikipedia articles. 
However, as these approaches do not incorporate truth-related concept, the probe may be misled to discover directions biased towards named entities, for instance. This limits generalizability to other datasets. 

\section{Conclusion}

In this work, we introduce IRIS, a novel unsupervised hallucination detection framework utilizing the internal states of LLMs. Specifically, an LLM is prompted to carefully verify the veracity of a given statement, and the contextualized embeddings of its response are obtained. 
Then, the uncertainty of the answer is evaluated, and regarded as a soft label for the truthfulness of the statement. Finally, a classifier is trained on the embeddings and corresponding pseudolabels. IRIS demonstrates state-of-the-arts performance on recent hallucination benchmarks, improving over strong baselines by a large margin. Our method does not incur much computational overhead and does not require enormous training data, making it suitable for real-time detection. 

Further investigation, using SAR scores as pseudolabels, shows the potential of incorporating more advanced uncertainty metrics into the IRIS pipeline (see Appendix~\ref{subsec:pseudo}). Other uncertainty approaches that are not as computationally costly can be explored. A preliminary analysis reveals that performance depends on the layer depth of the embeddings. More complex probe architecture can be exploited to aggregate embeddings at different layers. 
In addition, evaluation of embeddings at different tokens may uncover more granular insights into how internal representations correlate with truthfulness.

\section*{Limitations}

We believe our work has the following limitations:

\paragraph{Model Variety.} This work uses instruction-tuned models, ranging from 500 million parameters to 32 billion parameters. A wider family of models and of different sizes could provide additional insights.

\paragraph{Data Coverage.} The datasets used comprised single statements with clear-cut truth values for checking. In practice, we are more keen on checking entire passages, with a combination of factual statements that require verification, and non-factual claims that do not. A comprehensive system to effectively decompose the passages and filter statements before hallucination detection would make our pipeline more complete.

\paragraph{Interpretability of Internal Signals.} Although our method benefits from accessing latent internal activations, the black-box nature of these signals poses challenges for interpretability. Understanding the exact relationship between these internal cues and the truthfulness of a statement remains an open problem, suggesting that future work should also focus on enhancing the transparency and explainability of the detection process.

\section*{Acknowledgments}

We thank all the anonymous reviewers for their constructive feedback on our work. This research is supported by DSO grant DSOCL23216.

\bibliography{custom}

\clearpage

\appendix

\section*{Appendix}

\section{Prompt Templates}
\label{app:prompt}
\begin{figure}[h]
    \centering
    \includegraphics[width=\linewidth]{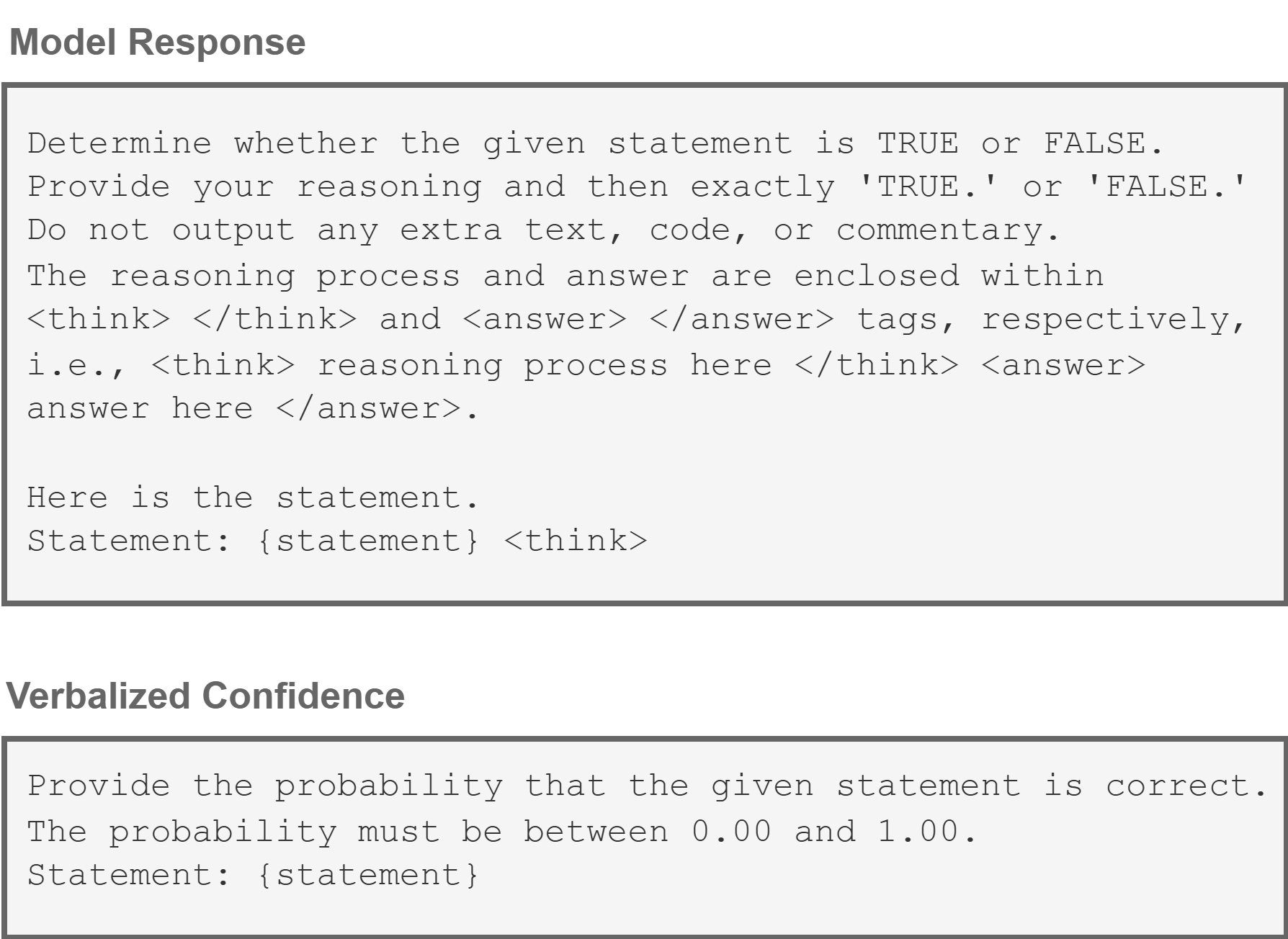}
    \caption{Prompts to verify statement correctness and to obtain verbalized confidence.}
    \label{fig:prompt}
\end{figure}

\section{Prompt Sensitivity}
\label{app:sensitivity}

\rev{To test sensitivity to different prompts, we consider 3 additional common prompting approaches on the True-False dataset. The first one is zero-shot CoT prompt (CoT-0). The other two are adversarial prompts, as suggested by \citet{turpin2023language}: (i) Suggested Answer: we append "I think the answer is false" to the end of every query; and (ii) Always False: in the few-shot demonstration, the answer is always "False",  regardless of the reason.}

\begin{table}[htbp]
    \centering
    \setlength{\tabcolsep}{1mm}
    \resizebox{\linewidth}{!}{
    {\color{black}
    \begin{tabular}{lccccccc|c} 
        \toprule
        \textbf{Prompt} & \textbf{Anim.} & \textbf{Cit.} & \textbf{Comp.} & \textbf{Elem.} & \textbf{Facts} & \textbf{Gen.} & \textbf{Invent.} & \textbf{Avg}  \\
        \midrule
        \textbf{CoT-0} & 82.18 & 92.12 & 94.58 & 90.86 & 95.93 & 83.67 & 94.32 & 91.16 \\
        \textbf{Suggested} & 79.70 & 91.78 & 92.50 & 89.78 & 91.87 & 86.36 & 87.50 & 88.91 \\
        \textbf{Always F} & 82.18 & 92.81 & 93.33 & 87.10 & 94.31 & 79.59 & 77.27 & 87.86 \\
        \textbf{CoT}       & 80.20 & 93.84 & 94.17 & 89.25 &  93.50 & 87.76 & 90.91 & 90.38 \\
        \bottomrule
    \end{tabular}
    }}
\end{table}

\rev{A slight drop in performance is observed for the adversarial prompts, but overall, IRIS is robust to these prompting techniques. With the adversarial prompts, the response is mostly correct, but the final answer is forced to comply with the demonstrations, resulting in a wrong answer. However, the hidden states contain useful information of correctness, and thus, the classifier can correctly identify using the hidden states. On the other hand, when these adversarial prompts are applied to get the verbalized confidence, the scores are all pushed to zero, rendering the pseudolabels ineffective for training the classifier probe.}

\section{Ablation Study}
\label{sec:ablation}

\begin{figure}
    \centering
    \includegraphics[width=\linewidth]{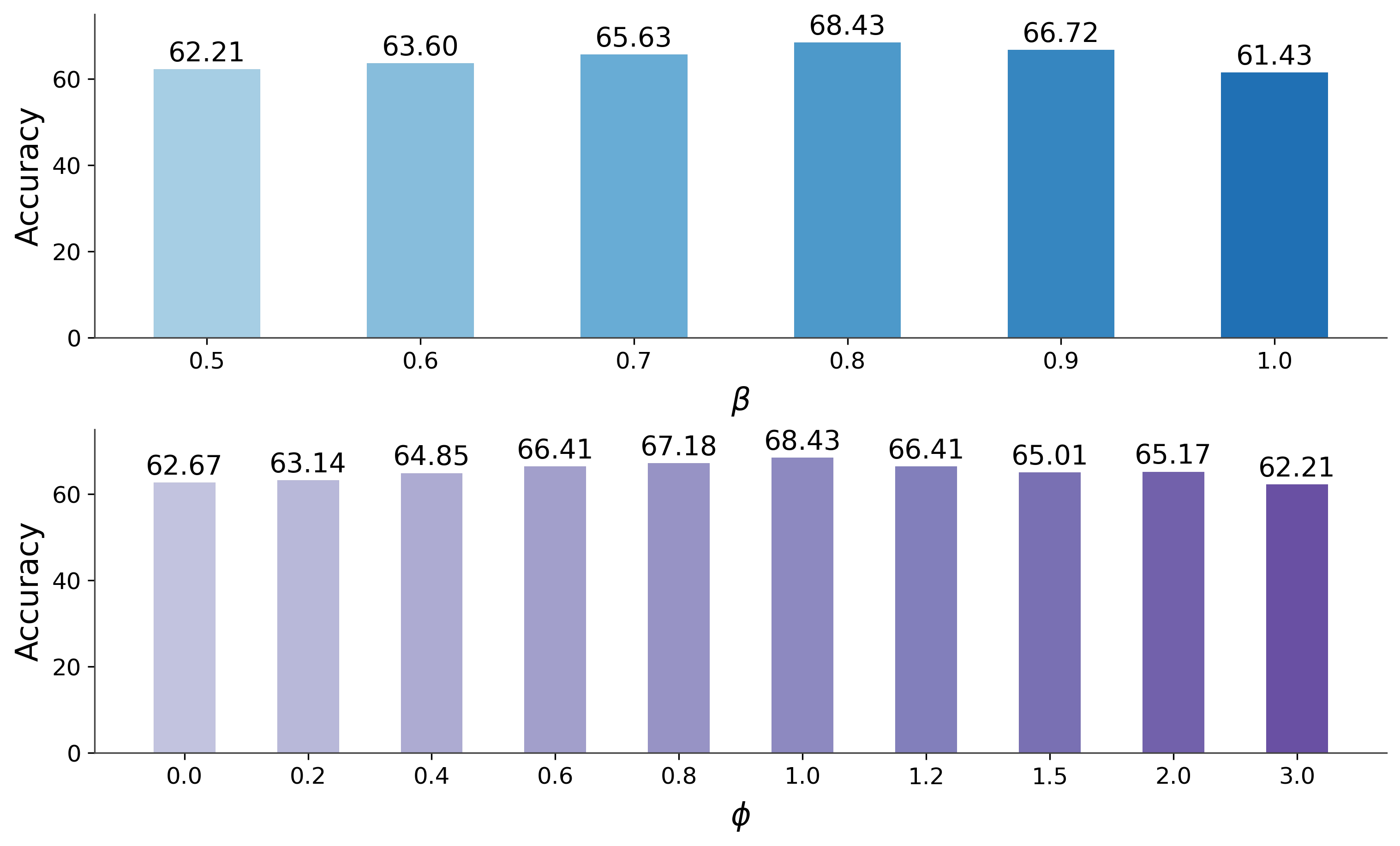}
    \caption{HELM average accuracy (\%) with Llama-3.1-8B-Instruct for various values of $\beta$ and $\phi$.}
    \label{fig:finetune}
\end{figure}

\paragraph{Classifier Hyperparameters.} In \Cref{fig:finetune}, we provide results for various values of the hyperparameters $\beta$ and $\phi$. $\beta$ denotes the importance given to the pseudolabel $\tilde{y}$ compared to the classifier's current state $\hat{y}$. $\beta$ cannot be too low and requires careful tuning. Meanwhile, the accuracy is less sensitive to $\phi$, except for when $\phi>1$ where a considerable drop in performance is observed. Overall, the incorporation of soft bootstrapping ($\beta<1$) and the symmetric cross entropy loss ($\phi>0$) provides additional performance gains.

\paragraph{Loss Function.} We further tested using $L_1$ and $L_2$ loss to train the classifier. The bootstrapping mechanism is maintained with $\beta=0.8$. The accuracy with $L_1$ and $L_2$ loss are 62.51\% and 58.16\%, respectively. A significant drop in accuracy is observed, highlighting the benefit of using the cross entropy loss.

\section{Case Study}

\rev{While CoT prompting generally improves performance for large models, smaller models do not always enjoy the same benefits. Here, we provide some examples from Llama-3.1-8B-Instruct when reasoning is helpful and when it is harmful.}

\rev{\paragraph{Good case.} For the statement \texttt{\small{'Human uses for hyena include conservation, research.'}}, the reasoning is provided as: \texttt{\small{'Hyenas are primarily viewed as pests or scavengers and are not typically kept in captivity for conservation or research purposes. </think>\textbackslash n\textbackslash n<answer> FALSE. </answer>'}} 
Without reasoning, the model incorrectly claims the statement is true.}

\rev{\paragraph{Bad case.} For \texttt{\small{'The giant anteater uses walking for locomotion.'}}, even though it is true, the reasoning is provided as: \texttt{\small{'The giant anteater primarily uses its powerful front legs and long claws for walking, but it also uses its long, sharp claws for digging and its tail for balance. </think>\textbackslash n\textbackslash n<answer> FALSE. </answer>'}}. 
Although the intermediate reasoning is correct, it deduced wrongly that the statement is false. 
However, its internal states reflect more accurately what the model knows, and the probe accordingly classifies the statement as correct.}

\section{Future Extensions}

\subsection*{Other Uncertainty Scores as Pseudolabels}
\label{subsec:pseudo}
\rev{IRIS offers a bridge between uncertainty-based methods and internal knowledge extraction. More advanced uncertainty metrics, such as SAR \citep{duan-etal-2024-shifting}, can be flexibly incorporated to generate pseudolabels to further enhance IRIS. The performance is shown in \Cref{fig:pseudo}. Compared to simply using SAR as a threshold (see \Cref{tab:main_results}), IRIS efficiently extracts the information embedded in the internal states to accurately assess the statement veracity.}

\begin{table}[htbp]

    \begin{subtable}{.5\textwidth}
        \centering
        \setlength{\tabcolsep}{1.5mm}
        \resizebox{\linewidth}{!}{
        \begin{tabular}{lccccccc|c}
            \toprule
            \textbf{Label} & \textbf{Anim.} & \textbf{Cit.} & \textbf{Co.} & \textbf{El.} & \textbf{Facts} & \textbf{Gen.} & \textbf{Inv.} & \textbf{Avg}  \\ 
            \midrule
            \textbf{SAR} & 70.79 & 73.29 & 78.25 & 59.68 & 85.37 & 79.59 & 77.84 & 73.88 \\
            \textbf{Verb} & 80.20 & 93.84 & 94.17 &  89.25 &  93.50 & 87.76 &  90.91 & 90.38 \\
            \bottomrule
        \end{tabular}
        }
        \caption{True-False}
    \end{subtable}
    
    \begin{subtable}{.5\textwidth}
        \centering
        \setlength{\tabcolsep}{2.8mm}
        \resizebox{\linewidth}{!}{
        \begin{tabular}{lccccc|c} 
            \toprule
            \textbf{Label} & \textbf{Bio-Med} & \textbf{Edu} & \textbf{Fin.} & \textbf{Open} & \textbf{Sci.} & \textbf{Avg} \\
            \midrule
            \textbf{SAR} & 67.47 & 68.03 & 77.25 & 93.22 & 76.50 & 74.38 \\
            \textbf{Verb} & 59.64 & 63.95 & 78.84 & 93.22 & 70.00 & 70.57 \\
            \bottomrule
        \end{tabular}
        }
        \caption{HaluEval2}
    \end{subtable}

    \begin{subtable}{.5\textwidth}
        \centering
        \setlength{\tabcolsep}{0.9mm}
        \resizebox{\linewidth}{!}{
        \begin{tabular}{lcccccc|c} 
            \toprule
            \textbf{Label} & \textbf{Falcon} & \textbf{GPT-J} & \textbf{LLB-7B} & \textbf{LLC-7B} & \textbf{LLC-13B} & \textbf{OPT-7B} & \textbf{Avg} \\
            \midrule
            \textbf{SAR}       & 65.71 & 61.74 & 68.14 & 63.00 & 67.71 & 70.18 & 66.10 \\ 
            \textbf{Verb}       & 67.62 & 69.57 &  69.03 & 66.00 & 69.79 & 68.42 & 68.43 \\ 
            \bottomrule
        \end{tabular}
        }
        \caption{HELM}
    \end{subtable}
    \caption{SAR uncertainty as pseudolabels.}
    \label{fig:pseudo}
\end{table}

\subsection*{Specialized Dataset}

Our preliminary investigation with mathematical reasoning datasets, i.e Arithmetic and GSM8K \citep{cobbe2021training}, shows that IRIS can potentially tackle specialized datasets. In particular, IRIS is applied to identify correct or incorrect answers to mathematical questions. IRIS achieves an accuracy of 87.2\% and 73.4\% on the Arithmetic and GSM8K datasets, respectively, compared to CoT prompting with 80.6\% and 69.9\%. However, on these datasets, it is more challenging to elicit verbalized confidence. When the prompting is done poorly, we observe that the model outputs a confidence of 0 for most queries, making the pseudolabels misguiding. In future work, experiments can be conducted on a wider range of reasoning datasets to understand the effectiveness of IRIS on these specialized domains.

\end{document}